\documentclass[journal]{IEEEtran}
\usepackage{amssymb} 
\usepackage{autobreak} 
\usepackage{booktabs}
\usepackage{cite}
\usepackage{marvosym}
\usepackage{times}
\usepackage{epsfig}
\usepackage{graphicx}
\usepackage{amssymb}
\usepackage{booktabs}
\usepackage{subcaption}
\usepackage{xcolor}
\usepackage{multirow}
\usepackage{graphics}
\usepackage{caption}
\usepackage{colortbl} 

\ifCLASSINFOpdf
\else
   \usepackage[dvips]{graphicx}
\fi
\usepackage{url}

\usepackage{graphicx}

\begin{document}

\title{Learning with Noisy Low-Cost MOS for Image Quality Assessment via Dual-Bias Calibration}

\author{Lei Wang, Qingbo Wu, \IEEEmembership{Member, IEEE}, Desen Yuan, King Ngi Ngan, \IEEEmembership{Life Fellow, IEEE}, Hongliang Li, \IEEEmembership{Senior Member, IEEE}, Fanman Meng, \IEEEmembership{Member, IEEE} and Linfeng Xu  
\thanks{The authors are with the School of Information and Communication Engineering, University of Electronic Science and Technology of China, Chengdu, 611731, China.}
}

\markboth{Journal of \LaTeX\ Class Files, Vol. 14, No. 8, August 2015}
{Shell \MakeLowercase{\textit{et al.}}: Bare Demo of IEEEtran.cls for IEEE Journals}
\maketitle

\begin{abstract}
Learning based image quality assessment (IQA) models have obtained impressive performance with the help of reliable subjective quality labels, where mean opinion score (MOS) is the most popular choice. However, in view of the subjective bias of individual annotators, the labor-abundant MOS (LA-MOS) typically requires a large collection of opinion scores from multiple annotators for each image, which significantly increases the learning cost. In this paper, we aim to learn robust IQA models from low-cost MOS (LC-MOS), which only requires very few opinion scores or even a single opinion score for each image. More specifically, we consider the LC-MOS as the noisy observation of LA-MOS and enforce the IQA model learned from LC-MOS to approach the unbiased estimation of LA-MOS. In this way, we represent the subjective bias between LC-MOS and LA-MOS, and the model bias between IQA predictions learned from LC-MOS and LA-MOS (i.e., dual-bias) as two latent variables with unknown parameters. By means of the expectation-maximization based alternating optimization, we can jointly estimate the parameters of the dual-bias, which suppresses the misleading of LC-MOS via a gated dual-bias calibration (GDBC) module. To the best of our knowledge, this is the first exploration of robust IQA model learning from noisy low-cost labels. Theoretical analysis and extensive experiments on four popular IQA datasets show that the proposed method is robust toward different bias rates and annotation numbers and significantly outperforms the other learning based IQA models when only LC-MOS is available. Furthermore, we also achieve comparable performance with respect to the other models learned with LA-MOS. 
\end{abstract}


\IEEEpeerreviewmaketitle

\section{Introduction}
\label{sec:intro}
Image quality assessment (IQA) is an active research area in multimedia technology, which is critical for evaluating and developing various perceptual-friendly image/video applications ~\cite{zhai2020perceptual,deng2017image,lin2019kadid,kundu2017large,zaric2011vcl,song2022knowledge,wang2020blind}. With the rapid development of the deep neural network (DNN), learning based IQA models are receiving more and more attention. Various advanced networks have been developed for IQA and achieved impressive performance recently. It is important to notice that the success of learning based IQA models highly relies on reliable subjective quality labels, which determine the direction of optimizing these advanced networks. Unfortunately, due to the subjective bias of individual annotators, it is greatly challenging to collect reliable subjective quality labels, especially on a large-scale image database. 
 
\begin{figure} 
    \centering
    \subfloat[]{
    \begin{minipage}[t]{0.46\linewidth}
    \centering
    \includegraphics[width=1.0\linewidth]{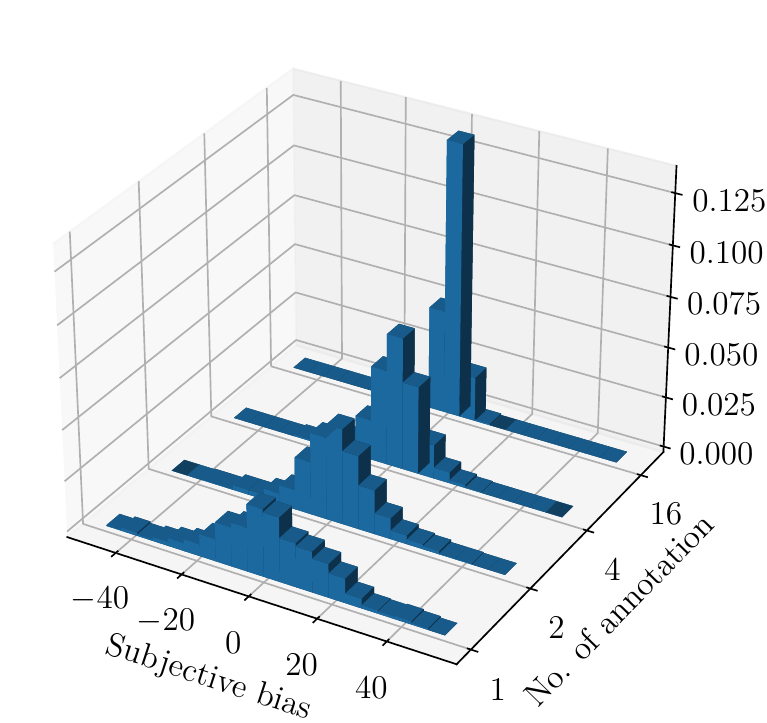}
    \end{minipage}
    } 
    \subfloat[]{
    \begin{minipage}[t]{0.46\linewidth}
    \centering
    \includegraphics[width=1.0\linewidth]{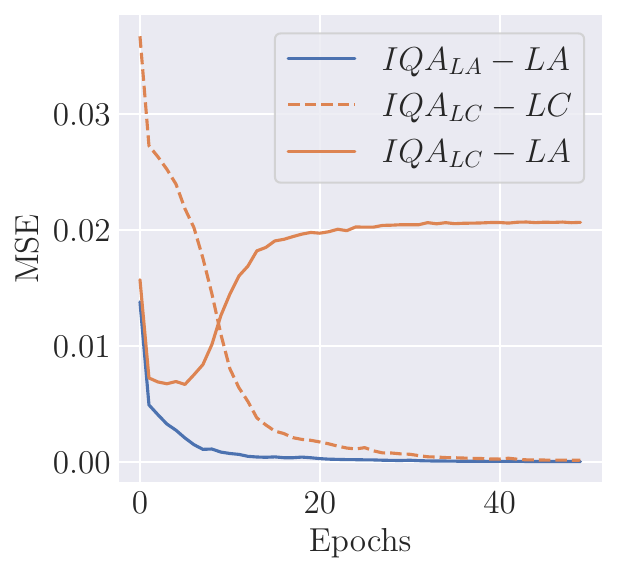}
    \end{minipage}    
    }
    \caption{(a) The distributions of the subjective bias under different numbers of human annotations on the VCL dataset; (b) The loss curves of training the DBCNN model with different labels on the KONIQ dataset.}
    \label{fig:case}
    \vspace{-0.9em}
\end{figure}

As the most widely used label for IQA in ideal conditions, the typical mean opinion score in existing datasets is labor-abundant (LA-MOS), which requires a large collection of opinion scores from multiple annotators (usually more than 15) for each image. 
Under a strictly standardized subjective experiment, the mean value of all annotators' opinion scores is finally used to represent their majority decision, which attempts to eliminate the subjective bias from individual annotators. 
As shown in Fig. \ref{fig:case} (a), subjective bias is presented as a Gaussian distribution, and the variance of the distribution diminishes as the annotation number increases. Meanwhile, different methods for pre/post-screening of the subject are also developed in BT500~\cite{bt2020methodologies}, P910~\cite{installationsitu}, and P913~\cite{union2016methods}, to further refine the regular LA-MOS by rejecting or weakening the votes of a subject, whose behavior is biased and inconsistent with the others. 
Despite efficient subjective bias reduction, the aforementioned methods are only applicable to LA-MOS, whose annotators are large enough for each image. On one hand, this label collection process is time-consuming and expensive, which significantly increases the learning cost. On the other hand, the model bias caused by subjective bias and the corresponding calibration strategy are both rarely investigated in previous works.

In this paper, we aim to learn robust IQA models from low-cost MOS (LC-MOS), which only requires very few opinion scores or even a single opinion score for each image.
LC-MOS exists in practical scenes and its corresponding noise form is subjective bias presented as a Gaussian distribution, which is different from the uniform distribution of the classification noisy label scene. 
Then we explore the negative impact of LCMOS on learning-based IQA models.
As shown in Fig. \ref{fig:case} (b), "$IQA_{LA}-LA$" represents the MSE curve between the output of the model trained by LA-MOS and its fit target LA-MOS; 
"$IQA_{LC}-LC$" represents the MSE curve between the output of the model trained with LC-MOS and its fit target LC-MOS; "$IQA_{LC}-LA$" represents the MSE curve between the output of the model trained with LC-MOS and its potential fit target LA-MOS.
Although "$IQA_{LC}-LC$"  (the orange dotted line) can reach 0 as well as "$IQA_{LA}-LA$"  (the blue line), "$IQA_{LC}-LA$" deviates from 0 (the orange solid line), which means that the IQA model prediction trained with LC-MOS is far from its real potential fitting target LA-MOS.

To address this issue, we consider the LC-MOS as the noisy observation of LA-MOS and enforce the IQA model learned from LC-MOS to approach the unbiased estimation of LA-MOS. In this way, we represent the dual-bias, including the subjective bias between LC-MOS and LA-MOS and model bias between IQA models learned from LC-MOS and LA-MOS, as two latent variables with unknown parameters. By means of the expectation-maximization based alternative optimization, we can jointly estimate the parameters of the dual-bias and adaptively suppress the misleading of LC-MOS via a gated dual-bias calibration (GDBC) module. 
Meanwhile, GDBC also achieved comparable performance in terms of four metrics using LC-MOS to the IQA model learned from LA-MOS. To the best of our knowledge, this is the first exploration of robust IQA model learning from noisy LC-MOS. For clarity, the main contributions of this paper are summarized in the following:
\begin{itemize}
    \item We propose an alternative optimization based dual-bias (including subjective bias and model bias) calibration method for robust IQA model learning from noisy LC-MOS, which significantly reduces the learning cost.
    \item We develop a GDBC module to adaptively update the estimated subjective bias by measuring the stability of IQA model learning in neighboring iterations, which reduces the risk of overadjustment.
    \item We verify the effectiveness of the proposed GDBC method from both the theoretical and experimental analysis, which achieves state-of-the-art performance when very few opinion scores are available for each image.
\end{itemize}

\section{Related Work}
In this section, we first briefly review learning based IQA methods, then introduce the related works about robust classification models and subjective annotation of practical conditions.

\subsection{Learning based Image Quality Assessment} 
As an important research area in image processing, IQA is essential in many applications such as image compression, image restoration, medical imaging, and etc, where the quality of the visual information is critical for accurate analysis and interpretation.
Recently, learning based IQA has received considerable attention due to the powerful ability of DNNs.
For example, 
Ma et al. \cite{ma2017end} proposed a multi-task DNN for IQA with learning based end-to-end optimization utilizing auxiliary distortion information.
Zhang et al. \cite{zhang2018blind} designed a deep bilinear convolutional neural network (DBCNN) for IQA with both synthetic and authentic distortions.
Talebi et al. \cite{talebi2018nima} proposed a convolutional neural network predicting score distribution for neural image assessment (NIMA).
Su et al. \cite{su2020blindly} proposed a self-adaptive hyper-network architecture for IQA (HyperIQA) in the wild.
Zhang et al. \cite{zhang2021uncertainty} proposed a unified IQA model and an approach training for both synthetic and realistic distortions.
Sun et al. \cite{sun2022graphiqa} developed a distortion graph representation learning framework for IQA. 
In addition, new paradigms for learning based IQA have emerged in complex scenarios. For example, the meta-learning IQA for fast adaptation \cite{zhu2020metaiqa}, the generative adversarial network (GAN) for active inference \cite{ma2021blind}, the evolvable predictive head for continuous learning \cite{zhang2022continual}, the disentangled representation based on variational auto-encoders (VAE) for image generation \cite{wang2020blind}, the vision-language correspondence \cite{zhang2023blind} for multimodal scenarios, and the perceptual attack for security scenarios \cite{zhang2022perceptual}. 

Since deep learning is an end-to-end optimization that relies on quality score regression, the superior performance of the above learning based IQA models heavily depends on reliable subjective quality labels, especially on a large-scale image database. 
Moreover, there is still a strong demand to construct new datasets for many emerging scenarios and tasks, including distorted images~\cite{lin2019kadid}, virtual reality (VR)~\cite{duan2018perceptual,sun2019mc360iqa}, light field~\cite{Vamsi2017}, hazy images~\cite{min2018objective,min2019quality}, smartphone photography~\cite{fang2020perceptual} and etc. 
Therefore, it is necessary and urgent to explore feasible IQA models in the subjective bias scenario.

\subsection{Robust Model Learning and Subjective Label Screening}
Completely clean labels are difficult to obtain in practical conditions. 
Researchers have found the over-parameterized network can learn any complex function from corrupted labels \cite{arpit2017closer,ghosh2017robust,tanaka2018joint,goldberger2016training,goldberger2016training,xia2021sample}. Zhang et al. \cite{zhang2021understanding} demonstrated that DNNs can easily fit the entire training dataset with any corrupted label ratio, ultimately leading to less generality on the test dataset. 
To train efficient DNNs in noisy cases, many methods have been proposed including robust loss functions~\cite{ghosh2017robust,tanaka2018joint,zhang2018generalized}, regularization~\cite{zhou2021learning}, 
robust network architecture~\cite{goldberger2016training}, 
sample selection~\cite{xia2021sample}, 
training strategy~\cite{hahn2019self,gotmare2018closer}, and etc.  
These methods focus on robust classification problems and the noise of perturbed labels is assumed to obey a uniform distribution. The design of most methods is based on noise tolerance and one-hot label properties such as sparsity regularization~\cite{wang2019symmetric, zhou2021learning}. 
These methods cannot be directly transferred to LC-MOS for IQA regression tasks due to the inconsistency of data properties.

In the practical condition for IQA, the opinion score of each annotator is biased against the ideal objective label, which is different from artificial perturbation in classification problems~\cite{zhuang2015leveraging,hube2019understanding,li2016crowdsourced}. There are different acquisition and processing methods for IQA datasets \cite{lin2019kadid}, such as ensuring the consistency of the subjective evaluation environment \cite{kundu2017large}, adding post-processing to the collected data \cite{virtanen2014cid2013}, and discarding outliers \cite{zaric2011vcl}. These datasets are collected through crowdsourcing and require multiple annotators' opinion scores, which is very time-consuming and expensive. 
Furthermore, the International Telecommunication Union (ITU) and researchers have proposed a number of standards~\cite{bt2020methodologies,installationsitu,union2016methods} for crowd-sourced data processing to eliminate the subjective bias in MOS, containing the model based on subject rejection in BT500~\cite{bt2020methodologies}, 
the model based on subject bias/inconsistency modeling and maximum likelihood estimation in P910~\cite{installationsitu,li2020simple},
and the model based on subject bias removal in P913~\cite{union2016methods}. 
However, these methods rely on sufficient annotation information and cannot be directly transferred to IQA models under LC-MOS scenarios.

\section{Methods}\label{sec:pl}
In this section, we first introduce some preliminary of subjective bias problem formulation. Then, we describe the misleading effect of LC-MOS, and then propose an expectation-maximization based dual-bias calibration scheme. 
\subsection{Preliminaries}
Let $x \in \mathcal{X} \subset \mathbb{R}^d$ denote an $d$-dimensional image, and $y/y^*\in \mathcal{Y}$ denote its corresponding LC-MOS/LA-MOS, i.e.,
\begin{equation}
\begin{split}
    &y=\frac{1}{M}\sum_{m=1}^M R_m\\
    &y^*=\frac{1}{S}\sum_{s=1}^S R_s
\end{split}, 
\end{equation}
where $R_m$/$R_s$ denote the $m$ th/$s$ th manual annotation for $x$, $\{R_m\}_{m=1}^M\subset \{R_s\}_{s=1}^S$, and $M\ll S$. To simplify the discussion, we represent the LC-MOS and LA-MOS with a normalized label space, i.e., $\mathcal{Y}\subset[0,1]$, and a higher $y/y^*$ means better subjective quality in terms of very few/abundant manual annotations. In this context, the subjective bias $z$ is defined as the difference between $y$ and $y^*$, i.e.,
\begin{equation}
    z=y-y^*,
    \label{eq-z}
\end{equation}
which is assumed to follow a Gaussian distribution with unknown parameters, i.e., $z\sim\mathcal{N}\left(\mu_z, \sigma_{z}^{2}\right)$.

Learning based IQA aims to obtain a parametric model $f_\theta: \mathcal{X} \rightarrow \mathcal{Y}$ that maps the image space into the subjective quality based label space, where $\theta$ is learned from paired image and label samples. Typically, the LA-MOS serves as the label, and we
derive the optimal parameter $\theta^{*}$ by minimizing the risk $R$ defined in the following, 
\begin{equation}
\theta^*=\arg\min_{\theta\in\Theta}R(\theta),
\end{equation}
where $\Theta$ is the available parameter set for $f_\theta$, and $R(\theta)$ is measured with the expectation of the loss between $f_\theta(x)$ and $y^*$ on the training set, i.e.,
\begin{equation}
R(\theta)=\mathbb{E}_{x,y^*}[\mathcal{L}(f_{\theta}(x),y^*)],
\end{equation}
where $\mathcal{L}(\cdot,\cdot)$ is the loss of the IQA model with respect to the label for each training sample. To save the annotation costs, this paper tries to replace parts of LA-MOS with LC-MOS, which may mislead the IQA model learning. Let $\theta^*$ denote the optimal parameters learned from $y^*$. We define the model bias $b$ by
\begin{equation}
    b=f_{\theta}(x)-f_{\theta^*}(x).
    \label{eq-b}
\end{equation}

In the following, a theoretical analysis of LC-MOS's misleading effect would be conducted on the popular square error loss function. Then, we put forward our gated dual-bias calibration (GDBC) method to efficiently suppress the aforementioned misleading effect, which enforces $f_{\theta}(x)$ to approach $y^*$ rather than $y$.

\subsection{Misleading Effect of LC-MOS}
Following the discussion in \cite{ghosh2017robust}, we denote the noisy labels with bias rate $\eta$ by $y^{\eta}$, i.e.,
\begin{equation}
    y^{\eta}=\left\{
    \begin{array}{ll}
        y, &\text{with probability}~\eta   \\
        y^*, &\text{with probability}~1-\eta  
    \end{array}.
    \right.
\end{equation}
Then, given a collection of training samples with previous noisy labels $y^{\eta}$, IQA model learning usually employs square error to measure the loss of $f_{\theta}(x)$ with respect to $y^{\eta}$, i.e.,
\begin{equation}
    \mathcal{L}\left(f_{\theta}(x),y^{\eta}\right)=\|f_{\theta}(x)-y^{\eta}\|_2^2.
    \label{eq-loss}
\end{equation}
Let $R^{\eta}(\theta)=\mathbb{E}_{x, y^{\eta}}[\mathcal{L}(f_{\theta}(x),y^{\eta})]$ denote the risk with bias rate $\eta$, and $\theta^{*,\eta}$ denote the parameter for the global minimum of risk $R^{\eta}(\cdot)$. To facilitate the analysis, we rewrite the risk $R^{\eta}(\theta)$ with bias rate $\eta$ in the following expanded form, i.e.,
\begin{equation} 
	\begin{aligned}  
    		R^{\eta}(\theta)=&\mathbb{E}_{{x},y^{\eta}}[\mathcal{L}(f_{\theta}({x}), y^{\eta})]\\ =&\mathbb{E}_{{x}} \mathbb{E}_{y^* \mid {x}} \mathbb{E}_{y^{\eta} \mid {x}, y^*}\left[ \mathcal{L}(f_{\theta}({x}), y^{\eta})\right] \\ 
    		=&\mathbb{E}_{{x}} \mathbb{E}_{y^* \mid {x}}\left\{ (1-\eta) \mathcal{L}(f_{\theta}({x}), y^*)+\right.\\
      & \eta \mathbb{E}_{z \mid {x}, y^*} [\mathcal{L}(f_{\theta}({x}), y^*+ z)]   \} \\ 
    		=&\mathbb{E}_{{x}} \mathbb{E}_{y^* \mid {x}} \left\{ (1-\eta) \mathcal{L}(f_{\theta}({x}), y^*)+ \right.\\ &
    		 \eta \mathbb{E}_{z \mid {x}, y^*} 
    		[\mathcal{L}(f_{\theta}({x}), y^*)-
    		2 z (f_{\theta}({x})-y^*)+ z^2] \} \\ 
    		=&\mathbb{E}_{{x}} \mathbb{E}_{y^* \mid {x}} \left\{
    		\mathcal{L}(f_{\theta}({x}), y^*)+\right.\\
      & \eta  \mathbb{E}_{z \mid {x}, y^*} [z^2-2z(f_{\theta}(x)-y^*)]   \}\\ 
    		=&R(\theta) -\eta\mathbb{E}_{{x}} \mathbb{E}_{y^* \mid {x}}     \left[2\mu_z(f_{\theta}(x)-y^*)+ \mu_z^2 +\sigma_{z}^{2} \right].
	\end{aligned} 
\end{equation}

Then, given the optimal and random parameters of $R(\cdot)$, i.e., $\theta^*$ and $\theta$, we can represent their risk difference $D$ under $R^{\eta}(\cdot)$ by
\begin{equation}
\begin{aligned}  
	D=&R^{\eta}\left(\theta^{*}\right)-R^{\eta}(\theta)\\
 =& R\left(\theta^{*}\right)-R(\theta)-\\&
 \eta\mathbb{E}_{{x}} \mathbb{E}_{y^* \mid {x}}     \left[2\mu_z(f_{\theta^{*}}(x)-y^*)+ \mu_z^2 +\sigma_{z}^{2} \right]+\\&
 \eta\mathbb{E}_{{x}} \mathbb{E}_{y^* \mid {x}}     \left[2\mu_z(f_{\theta}(x)-y^*)+ \mu_z^2 +\sigma_{z}^{2} \right]\\
 =&R\left(\theta^{*}\right)-R(\theta)+\eta\mathbb{E}_{{x}} \mathbb{E}_{y^* \mid {x}} [2\mu_zb].
\end{aligned} 	
\label{eq-risk-difference}
\end{equation}

Although $R(\theta^{*})-R(\theta)\leq 0$, we can not guarantee that $R^{\eta}\left(\theta^{*}\right)-R^{\eta}(\theta)\leq 0$ due to the uncertainty of $\mu_z$ and $b$, which are both probably nonzero and with the same signs. That is, $\theta^{*}$ does not necessarily equal to $\theta^{*,\eta}$ when training with the square error loss. In addition, this misleading effect of LC-MOS would become more significant when a higher bias rate $\eta$ or larger subjective bias $\mu_z$ are applied to the Eq. (\ref{eq-risk-difference}). Therefore, it is urgent to develop a robust learning framework to train the IQA model from the low-cost noisy labels.

\begin{figure*}
	\centering
	\includegraphics[width=.9\linewidth]{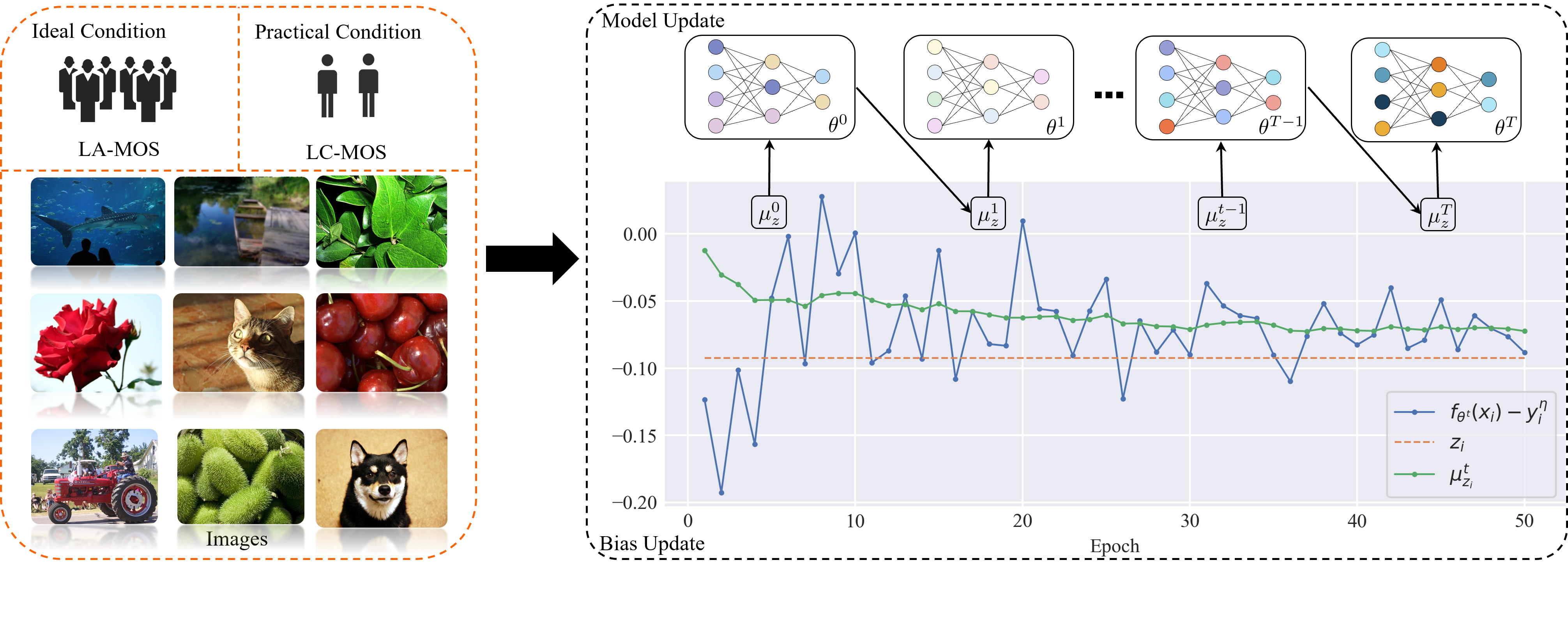}
	\caption{The framework of the proposed GDBC method for IQA. It alternates between model update and bias update through the EM algorithm, which is a process of mutual promotion.}
	\label{fig:arch}
\end{figure*}

\subsection{Expectation-maximization based Dual-bias Calibration}
We propose an expectation-maximization based dual-bias calibration to alleviate the above-mentioned challenges of biased LC-MOS. The framework of GDBC is shown in Fig. \ref{fig:arch}. The algorithm alternates between model update and bias update. During the model update process, the neural network parameters are updated through backpropagation, and image features are learned for quality evaluation. During the bias update process, we repeatedly estimate the subjective bias. The latter subjective bias in turn helps the model learn. The whole process is a process of mutual promotion so that the robustness of the model is improved.

According to Eq. (\ref{eq-risk-difference}), if we know $z$ and replace $y^{\eta}$ by $y^{\eta}-z$ in Eq. (\ref{eq-loss}), both the subjective and model biases would be pushed to zero, which improves the noisy label tolerance of IQA model learning. To this end, we consider $Z=\{(z_i)_{i=1}^n\}$ as the latent variable of $n$ training sample and represent the likelihood of the unknown parameter set $\Omega=\{\theta,(\omega_i)_{i=1}^n\}$ by
\begin{equation}
\begin{aligned}
    L(\Omega;Y^{\eta})=&\prod_{i=1}^n p(y_i^{\eta}|\theta,\omega_i)\\
\end{aligned},
\end{equation}
where $Y^{\eta}=\{(y_i^{\eta})_{i=1}^n\}$ denotes $n$ independently observed noisy labels and $\omega_i=\{\mu_{z_i},\sigma_{z_i}\}$. Following \cite{EM}, we employ the expectation-maximization based iterative method to derive $\Omega$ for maximizing $L(\Omega; Y^{\eta})$, which helps us achieve the dual-bias calibration.
Let $\Omega^t=\{\theta^t,(\omega_i^t)_{i=1}^n\}$ denote the estimated parameters in the $t$th iteration. We first conduct the Expectation step (E-step) by computing the conditional expectation of the log-likelihood, 

\begin{equation}
\begin{aligned}
Q(\Omega\mid \Omega^t)=&\mathbb{E} _{Z \mid Y^{\eta},\Omega^t}\left[ \log L(\Omega;Y^{\eta},Z)\right]\\
=&\mathbb{E} _{Z \mid Y^{\eta},\Omega^t}\left[ \log\prod_{i=1}^n p(y_i^{\eta},z_i\mid\Omega)\right]\\
=&\sum_{i=1}^n\int p(z_i\mid y_i^{\eta},\Omega^t)\left[\log p(y_i^{\eta}\mid z_i,\Omega)+\right.\\
&\left. \log p(z_i\mid \Omega)\right]dz_i
\end{aligned}.
\label{eq-cond-like}
\end{equation}
For dual-bias calibration, we want to use the calibrated label $y_i^{\eta}-z_i$ to supervise the IQA model $f_{\theta}(x_i)$, which could be transformed to maximizing the posterior of the observed noisy label $y_i^{\eta}$ with $f_{\theta}(x_i)+z_i$. Based on this requirement, we assume that ${y_i^{\eta}}_{\mid z_i,\Omega}\sim\mathcal{N}(f_{\theta}(x_i)+z_i,\sigma_{y_i^{\eta}\mid z_i,\Omega})$, which enforces $f_{\theta}(x_i)$ to approach $y_i^*$ according to Eq. (\ref{eq-z}). In addition, the conditional distribution of ${z_i}_{\mid y_i^{\eta},\Omega^t}$ could be derived from the Bayes theorem
\begin{equation}
    p(z_i\mid y_i^{\eta},\Omega^t)=\frac{p(y_i^{\eta}\mid z_i,\Omega^t)p(z_i\mid\Omega^t)}{\int p(y_i^{\eta}\mid z_i,\Omega^t)p(z_i\mid\Omega^t)dz_i},
\end{equation}
where we obtain\footnote{Detailed proof is given in the supplementary material.\label{proof}} that ${z_i}_{\mid y_i^{\eta},\Omega^t}\sim\mathcal{N}(\mu_{z_i\mid y_i^{\eta},\Omega^t},\sigma_{z_i\mid y_i^{\eta},\Omega^t})$ and
\begin{equation}
\label{uz}
    \mu_{z_i\mid y_i^{\eta},\Omega^t}=\frac{\sigma_{y_i^{\eta}\mid z_i,\Omega^t}^{2}\mu_{z_i}^{t}-\sigma_{z_i\mid  \Omega^t}^{2}[f_{\theta^{t}}(x_{i})-y_{i}^{\eta}]}{\sigma_{y_i^{\eta}\mid z_i,\Omega^t}^2+\sigma_{z_i\mid  \Omega^t}^2}.
\end{equation}

By plugging the probability density of ${z_i}_{\mid y_i^{\eta},\Omega^t}$, ${y_i^{\eta}}_{\mid z_i,\Omega}$ and ${z_i}_{\mid\Omega}$ into Eq. (\ref{eq-cond-like}), we rewrite the conditional expectation of the log-likelihood by
\begin{equation}
\begin{aligned}
    Q(\Omega\mid \Omega^t)=&-\frac{1}{2}\sum _{i=1}^{n} \left[ \frac{\sigma_{z_i\mid y_i^{\eta},\Omega^t}^2 + (\mu_{z_i} - \mu_{z_i\mid y_i^{\eta},\Omega^t})^2}{ \sigma_{z_i\mid  \Omega}^2}\right.\\
&\left. + \frac{\sigma_{z_i\mid y_i^{\eta},\Omega^t}^2 + [y_i-f_{\theta}(x)- \mu_{z_i\mid y_i^{\eta},\Omega^t}]^2}{ \sigma_{y_i^{\eta}\mid z_i,\Omega}^2}\right.\\ 
&\left.+\log (2 \pi \sigma_{z_i\mid  \Omega}^2)+\log (2 \pi \sigma_{y_i^{\eta}\mid z_i,\Omega}^2)  \right] 
\end{aligned}.
\end{equation}

Then, by setting the derivative of $Q(\Omega\mid \Omega^t)$ w.r.t. $\mu_{z_i}$ to zero, we could obtain the updated parameter of subjective bias  in the Maximization step (M step), i.e.,
\begin{equation}
    \mu_{z_i}^{t+1}=\alpha\mu_{z_i}^{t}+(1-\alpha)c_i^t,
    \label{eq-update}
\end{equation}
where $\alpha=\frac{\sigma_{y_i^{\eta}\mid z_i,\Omega}^{2}}{\sigma_{z_i\mid  \Omega}^2+\sigma_{y_i^{\eta}\mid z_i,\Omega}^2}$, $c_i^t=f_{\theta^t}(x_i)-y_i^{\eta}$, and $\mu_{z_i}^0=0$. It is noted that the fitting error $c_i^t$ usually converges to a small value when training with clean labels in several iterations \cite{fitting-error}. Unbounded updating of Eq. (\ref{eq-update}) may result in overadjustment. To address this issue, we develop a gated dual-bias calibration (GDBC) module by measuring the stability of IQA model learning in neighboring iterations, i.e.,
\begin{equation}\label{b}
      \mu_{z_{i}}^{t+1}=
      \begin{cases}
      \alpha \mu_{z_{i}}^{t}+(1-\alpha) c^{t}_{i}, & \| C\|_1 > t_h\epsilon\\
      \qquad  \ \mu_{z_{i}}^{t}, & \textit{otherwise}
     \end{cases},
\end{equation}
where $C=[c_i^{t-t_h},\cdots,c_i^{t}]^T$ represents the fitting errors of the IQA model in the neighboring $t_h$ iterations, and the subjective bias calibration only activates when $C$'s $l_1$ norm exceeds a threshold $\epsilon$.

Since $\mu_{z_i}^{t+1}$ maximizes the probability of $z_i$, we could suppress the model bias by removing $\mu_{z_i}^{t+1}$ from the noisy LC-MOS $y_i^{\eta}$, and optimize the IQA model parameters via
\begin{equation}
    \theta^{t+1}=\theta^{t}-\lambda\nabla_{\theta}\left[\frac{1}{n}\sum_{i=1}^n\mathcal{L}(f_{\theta^t}(x_i),y_i^{\eta}-\mu_{z_i}^{t+1})\right],
    \label{eq:final}
\end{equation}
where $\lambda$ is the learning rate, and $\nabla_{\theta}$ denotes the gradient operator.
We repeat the alternative subjective bias and model bias calibrations until the maximum iteration steps are reached. Finally, we could output a robust IQA model even when learning with noisy LC-MOS. 

\section{Experiments}\label{sec:exp}

\subsection{Protocol}
\label{sec:nmt}

We evaluate the proposed GDBC method on four popular IQA databases, i.e., VCL \cite{zaric2011vcl}, CSIQ \cite{ponomarenko2015image}, LIVEC \cite{ghadiyaram2015massive} and KONIQ \cite{hosu2020koniq}, which only provide the LA-MOS for all images. Let $M$ denote the annotation number to be simulated in the experiments. We develop the following three LC-MOS settings according to the annotation resources of different databases:

\begin{enumerate}
\item When raw opinion scores of all subjects are available (such as VCL \cite{zaric2011vcl}), we randomly sample $M$ scores for each image.
\item When pairwise MOS and standard deviation are available (such as CSIQ \cite{ponomarenko2015image} and LIVEC \cite{ghadiyaram2015massive}), we use them to simulate a normal distribution of subjective ratings \cite{wu2018perceptually,wu7937920}, and randomly sample $M$ scores from this distribution.
\item When the empirical distribution of raw opinion scores is available (such as KONIQ \cite{hosu2020koniq}), we randomly sample $M$ scores from this empirical distribution.
\end{enumerate}

For each image, the mean value of the previously sampled $M$ scores is used as the LC-MOS, where $M$ is smaller than the minimum requirement of ITU recommendations \cite{bt2020methodologies,installationsitu}. More specifically, we investigate four candidate numbers, i.e., $M=\{1, 2, 4, 8\}$. To validate the universality of the proposed method, we select four representative deep neural networks for IQA model learning, i.e., ResNet-50~\cite{he2016deep}, DBCNN~\cite{zhang2018blind},  HyperIQA~\cite{su2020blindly} and NIMA~\cite{talebi2018nima}. Following the criterion of \cite{zhang2018blind,su2020blindly,talebi2018nima}, we randomly split each database into non-overlapped training and testing sets, which cover 80\% and 20\% samples respectively. To eliminate the performance bias for specific LC-MOS or train-test split, we repeat the random LC-MOS sampling and train-test splitting 10 times for each database, and report the median results across all trials for evaluation. 

\begin{figure*}
    \centering
    \includegraphics[width=\linewidth]{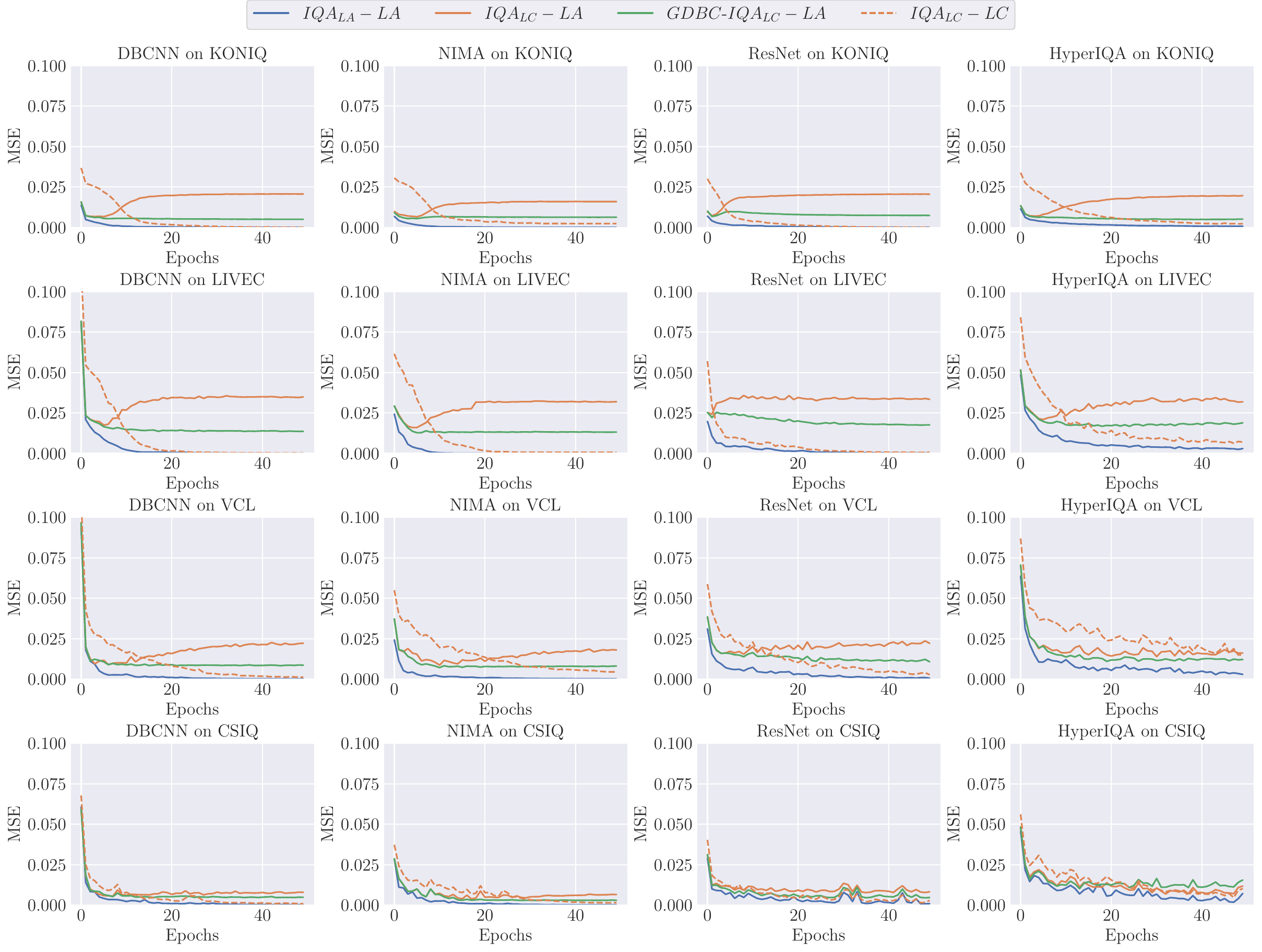} %
    \caption{The loss curves for different IQA models when training with LC-MOS and LA-MOS.}
    \label{fig:True MSE}
\end{figure*}

Let LA-MOS denote the ground-truth subjective quality of each image. Three widely used metrics are used for evaluating the performances of the IQA models learned from the LA-MOS, LC-MOS, and our GDBC method, i.e., the Pearson's Linear Correlation Coefficient (PLCC)~\cite{benesty2009pearson}, the Spearman's Rank Order Correlation Coefficient (SRCC)~\cite{zar2005spearman}, and Kendall’s Rank Correlation Coefficient (KRCC). In addition, to highlight the performance improvement of the proposed method when training with LC-MOS, we also introduce a relative index $\Delta(\%)$, i.e.,
\begin{equation}
    \Delta(\%)=\frac{m_{w/~GDBC}-m_{w/o~GDBC}}{m_{w/o~GDBC}}\times 100
\end{equation}
where $m_{w/~GDBC}$ and $m_{w/o~GDBC}$ denote the evaluation metric of an IQA model trained with and without the GDBC module, respectively.

In our experiment, we train the IQA models with Adam optimizer\cite{kingma2014adam} and set $\alpha$ to 0.9, $t_h$ to 1 or 3, $\epsilon$ to 0.01 or 0.1, epoch to $50$, batch size to $16$. The optimal learning rates are founded by grid search and scheduled by the cosine annealing rule\cite{loshchilov2016sgdr}. 
During training and inference, we scale and center crop $320\times 320\times 3$ sub-images from the original image without changing their aspect ratio. 
All experiments are performed on a workstation with a single NVIDIA GeForce RTX 3090 GPU.

\begin{table}[]
\centering
\caption{MSE between calibrated LC-MOS and LA-MOS} 
\begin{tabular}{c|ccccc}
\toprule
\midrule
Datasets & LC-MOS & DBCNN & HyperIQA & NIMA & ResNet \\ \midrule
KONIQ & 0.0207 & 0.0055 & \textbf{0.0046} & 0.0060 & 0.0076 \\
LIVEC & 0.0349 & \textbf{0.0139} & 0.0156 & 0.0139 & 0.0185 \\
VCL & 0.0239 & 0.0085 & 0.0098 & \textbf{0.0068} & 0.0121 \\
CSIQ & 0.0085 & 0.0045 & 0.0043 & \textbf{0.0030} & 0.0040 \\  
\midrule
\bottomrule
\end{tabular}
\label{tab:calibrated-MOS}
\end{table}

\subsection{Performance evaluation for the GDBC} 
To demonstrate the effectiveness of the proposed GDBC method, we first investigate the most challenging case with the setting of $\eta=100\%$ and $M=1$, which means that all training samples are labeled with the noisy LC-MOS and only one annotation is available for each image.

\subsubsection{Effectiveness of subjective bias calibration}
Let $y_i^\eta-\mu_{z_i}^{t+1}$ in Eq. \ref{eq:final} denote the calibrated LC-MOS. In Table \ref{tab:calibrated-MOS}, we report the subjective bias calibration results when GDBC is applied to different deep IQA models, which measure the mean square error (MSE) between the calibrated LC-MOS and its corresponding LA-MOS. Due to the difference in LC-MOS settings, the MSE between the raw LC-MOS and LA-MOS are different across different IQA databases, where the LIVEC and CSIQ present the largest and smallest bias respectively. Meanwhile, it is seen that the MSE values of the calibrated LC-MOS are much smaller than the raw LC-MOS when GDBC is applied to all deep IQA models, whose MSE reductions could range from 64\% to 77\%. These results verify the effectiveness and universality of the proposed method in reducing subjective bias. In the following, we will further investigate the benefits of subjective bias calibration for IQA model learning.

\subsubsection{Guidance for the IQA model learning} 
In Fig. \ref{fig:True MSE}, we show the loss curves of different deep IQA models when they are trained on multiple popular databases with different labels or training strategies.  
More specifically, $GDBC\mbox{-}IQA_{LC}-LA$ represents the MSE between an IQA model's output and the LA-MOS when LC-MOS and GDBC strategy are used for training; $IQA_{LA}-LA$ represents the MSE between an IQA model's output and the LA-MOS when LA-MOS is used for training; 
$IQA_{LC}-LC$ represents the MSE between an IQA model's output and the LC-MOS when LC-MOS is used for training; $IQA_{LC}-LA$ represents the MSE between an IQA model's output and the LA-MOS when LC-MOS is used for training. It is seen that both $IQA_{LA}-LA$ and $IQA_{LC}-LC$ could converge to a very small value except that the convergence speed of $IQA_{LC}-LC$ is slower than $IQA_{LA}-LA$, which is consistent with the observations of existing noisy label learning tasks \cite{fitting-error}. When we compare $IQA_{LC}-LC$ with $IQA_{LC}-LA$, it is found that the $IQA_{LC}$ presents a clear overfitting toward LC-MOS, whose loss quickly goes up with respect to LA-MOS after several epochs. Meanwhile, the overfitting issue of $IQA_{LC}-LC$ is very obvious for LIVEC but obsolete for CSIQ. Referring to Table \ref{tab:calibrated-MOS}, this observation confirms that a larger bias of raw LC-MOS would be more misleading for IQA model learning. By contrast, our $GDBC\mbox{-}IQA_{LC}-LA$ follows a more consistent tendency with $IQA_{LA}-LA$, which efficiently suppresses the rising of the loss with respect to LA-MOS across all IQA models and databases. This clearly shows the benefits of LC-MOS calibration for robust IQA model learning. 

\begin{table*}[] 
\centering
\caption{Performance of different methods on different datasets with biased labels (bias rate is $100\%$ and the annotation number is 1). $LA$ and $LC$ indicate training using LA-MOS and LC-MOS, respectively.}
\label{tab:table1}
\resizebox{2\columnwidth}{!}{
    \setlength{\tabcolsep}{1pt}
\begin{tabular}{l|rrr|rrr|rrr|rrr}
\toprule 
\toprule 
 
\multirow{2}{*}{Methods} & \multicolumn{3}{c|}{KONIQ} & \multicolumn{3}{c|}{LIVEC} & \multicolumn{3}{c|}{VCL} & \multicolumn{3}{c}{CSIQ} \\
 
 & SRCC $\uparrow$ & PLCC $\uparrow$ & KRCC $\uparrow$ & SRCC $\uparrow$ & PLCC $\uparrow$ & KRCC $\uparrow$ & SRCC $\uparrow$ & PLCC $\uparrow$ & KRCC $\uparrow$ & SRCC $\uparrow$ & PLCC $\uparrow$ & KRCC $\uparrow$ \\   \midrule  
\rowcolor[gray]{.8} $\text{ResNet}_{LA}$ & 0.8985 & 0.9214 & 0.7290 & 0.8215 & 0.8624 & 0.6327 & 0.9359 & 0.9234 & 0.7741 & 0.9242 & 0.9238 & 0.7575 \\
$\text{ResNet}_{LC}$ & 0.7905 & 0.8361 & 0.5990 & 0.7648 & 0.8021 & 0.5699 & 0.8503 & 0.8376 & 0.6469 & 0.8992 & 0.8961 & 0.7276 \\
GDBC-$\text{ResNet}_{LC}$ & 0.8294 & 0.8659 & 0.6425 & 0.7819 & 0.8179 & 0.5885 & 0.8723 & 0.8531 & 0.6724 & 0.9126 & 0.9224 & 0.7533 \\
$\Delta$(\%) & \textbf{+4.9209} & \textbf{+3.5642} & \textbf{+7.2621} & \textbf{+2.2359} & \textbf{+1.9698} & \textbf{+3.2637} & \textbf{+2.5873} & \textbf{+1.8505} & \textbf{+3.9419} & \textbf{+1.4902} & \textbf{+2.9349} & \textbf{+3.5322} \\ \midrule
\rowcolor[gray]{.8}$\text{NIMA}_{LA}$ & 0.8981 & 0.9179 & 0.7287 & 0.8011 & 0.8404 & 0.6112 & 0.9389 & 0.9067 & 0.7759 & 0.9263  & 0.9369  & 0.7661 \\
$\text{NIMA}_{LC}$ & 0.7551 & 0.7363 & 0.5626 & 0.6797 & 0.7017 & 0.4916 & 0.8846 & 0.8405 & 0.6987 & 0.9101 & 0.9222 & 0.7362 \\
GDBC-$\text{NIMA}_{LC}$ & 0.8216 & 0.8519 & 0.6357 & 0.7108 & 0.7376 & 0.5226 & 0.9325 & 0.9216 & 0.7754 & 0.9229 & 0.9349  & 0.7664  \\
$\Delta$(\%) & \textbf{+8.8068} & \textbf{+15.7001} & \textbf{+12.9932} & \textbf{+4.5755} & \textbf{+5.1161} & \textbf{+6.3059} & \textbf{+5.4149} & \textbf{+9.6490} & \textbf{+10.977} & \textbf{+1.4080} & \textbf{+1.3769} & \textbf{+4.0958} \\ \midrule
\rowcolor[gray]{.8}$\text{DBCNN}_{LA}$ & 0.8899 & 0.9083 & 0.7161 & 0.7390 & 0.7651 & 0.5615 & 0.9209 & 0.9199 & 0.7443 & 0.9498 & 0.9604 & 0.8096 \\
$\text{DBCNN}_{LC}$ & 0.7087 & 0.7474 & 0.5189 & 0.6579 & 0.6812 & 0.4756 & 0.8443 & 0.8633 & 0.6478 & 0.9233 & 0.9301 & 0.7600 \\
GDBC-$\text{DBCNN}_{LC}$ & 0.8053 & 0.8424 & 0.6137 & 0.7007 & 0.7095 & 0.5181 & 0.9074 & 0.9159 & 0.7377 & 0.9463 & 0.9552 & 0.8067 \\
$\Delta$(\%) & \textbf{+13.6306} & \textbf{+12.7107} & \textbf{+18.2694} & \textbf{+6.5055} & \textbf{+4.1544} & \textbf{+8.9361} & \textbf{+7.4736} & \textbf{+6.0929} & \textbf{+13.8777} & \textbf{+2.4903} & \textbf{+2.7033} & \textbf{+6.1507} \\ \midrule
\rowcolor[gray]{.8}$\text{HyperIQA}_{LA}$ & 0.8792 & 0.9104 & 0.7029 & 0.8183 & 0.8637 & 0.6387 & 0.9437 & 0.9279 & 0.7908 & 0.9191  & 0.9344  & 0.7523 \\
$\text{HyperIQA}_{LC}$ & 0.6709 & 0.7304 & 0.4868 & 0.6482 & 0.6943 & 0.4694 & 0.8734 & 0.8374 & 0.6842 & 0.8705 & 0.8760 & 0.6962 \\
GDBC-$\text{HyperIQA}_{LA}$ & 0.8012 & 0.8452 & 0.6109 & 0.7200 & 0.7564 & 0.5298 & 0.9390 & 0.9239 & 0.7860 & 0.9119  & 0.9348   & 0.7471 \\
$\Delta$(\%) & \textbf{+19.4217} & \textbf{+15.7174} & \textbf{+25.4930} & \textbf{+11.076} & \textbf{+8.9443} & \textbf{+12.867} & \textbf{+7.5109} & \textbf{+10.329} & \textbf{+14.878} & \textbf{+4.7504} & \textbf{+6.7099} & \textbf{+7.3149}   \\ \midrule 
\bottomrule
\end{tabular}
} 
\vspace{1em}
\end{table*}

\begin{table*}[] 
\centering
\caption{Performance of different methods with different bias rates. $LA$ and $LC$ indicate training using LA-MOS and LC-MOS, respectively.}
\label{tab:table2}
\resizebox{2\columnwidth}{!}{
    \setlength{\tabcolsep}{1pt}
\begin{tabular}{l|rrr|rrr|rrr|rrr}
\toprule 
\toprule 
\multirow{2}{*}{Methods} & \multicolumn{3}{c|}{$\eta=100\%$} & \multicolumn{3}{c|}{$\eta=80\%$} & \multicolumn{3}{c|}{$\eta=60\%$} & \multicolumn{3}{c}{$\eta=40\%$} \\
 & SRCC $\uparrow$ & PLCC $\uparrow$ & KRCC $\uparrow$ & SRCC $\uparrow$ & PLCC $\uparrow$ & KRCC $\uparrow$ & SRCC $\uparrow$ & PLCC $\uparrow$ & KRCC $\uparrow$ & SRCC $\uparrow$ & PLCC $\uparrow$ & KRCC $\uparrow$ \\ \midrule
\rowcolor[gray]{.8}$\text{ResNet}_{LA}$ & 0.8985 & 0.9214 & 0.7290 & 0.8985 & 0.9214 & 0.7290 & 0.8985 & 0.9214 & 0.7290 & 0.8985 & 0.9214 & 0.7290 \\
$\text{ResNet}_{LC}$ & 0.7905 & 0.8361 & 0.5990 & 0.8152 & 0.8572 & 0.6264 & 0.8385 & 0.8762 & 0.6536 & 0.8594 & 0.8946 & 0.6755 \\
GDBC-$\text{ResNet}_{LC}$ & 0.8294 & 0.8659 & 0.6425 & 0.8410 & 0.8778 & 0.6557 & 0.8593 & 0.8899 & 0.6785 & 0.8719 & 0.8994 & 0.6924 \\
$\Delta$(\%) & \textbf{+4.9209} & \textbf{+3.5642} & \textbf{+7.2621} & \textbf{+3.1649} & \textbf{+2.4032} & \textbf{+4.6775} & \textbf{+2.4806} & \textbf{+1.5636} & \textbf{+3.8097} & \textbf{+1.4545} & \textbf{+0.5366} & \textbf{+2.5019} \\ \midrule
\rowcolor[gray]{.8}$\text{NIMA}_{LA}$ & 0.8981 & 0.9179 & 0.7287 & 0.8981 & 0.9179 & 0.7287 & 0.8981 & 0.9179 & 0.7287 & 0.8981 & 0.9179 & 0.7287 \\
$\text{NIMA}_{LC}$ & 0.7551 & 0.7363 & 0.5626 & 0.7914 & 0.8301 & 0.5993 & 0.8099 & 0.8428 & 0.6185 & 0.8470 & 0.8767 & 0.6617 \\
GDBC-$\text{NIMA}_{LC}$ & 0.8216 & 0.8519 & 0.6357 & 0.8268 & 0.8645 & 0.6396 & 0.8376 & 0.8688 & 0.6517 & 0.8605 & 0.8890 & 0.6792 \\
$\Delta$(\%) & \textbf{+8.8068} & \textbf{+15.7001} & \textbf{+12.9932} & \textbf{+4.4682} & \textbf{+4.1444} & \textbf{+6.7335} & \textbf{+3.4202} & \textbf{+3.0850} & \textbf{+5.3678} & \textbf{+1.5939} & \textbf{+1.4030} & \textbf{+2.6447} \\ \midrule
\rowcolor[gray]{.8}$\text{DBCNN}_{LA}$ & 0.8899 & 0.9083 & 0.7161 & 0.8899 & 0.9083 & 0.7161 & 0.8899 & 0.9083 & 0.7161 & 0.8899 & 0.9083 & 0.7161 \\
$\text{DBCNN}_{LC}$ & 0.7087 & 0.7474 & 0.5189 & 0.7710 & 0.8191 & 0.5788 & 0.7954 & 0.8334 & 0.6033 & 0.8229 & 0.8553 & 0.6320 \\
GDBC-$\text{DBCNN}_{LC}$ & 0.8053 & 0.8424 & 0.6137 & 0.8318 & 0.8655 & 0.6431 & 0.8363 & 0.8676 & 0.6491 & 0.8514 & 0.8781 & 0.6663 \\
$\Delta$(\%) & \textbf{+13.6306} & \textbf{+12.7107} & \textbf{+18.2694} & \textbf{+7.8859} & \textbf{+5.6648} & \textbf{+11.1092} & \textbf{+5.1421} & \textbf{+4.1037} & \textbf{+7.5916} & \textbf{+3.4634} & \textbf{+2.6657} & \textbf{+5.4272} \\ \midrule
\rowcolor[gray]{.8}$\text{HyperIQA}_{LA}$ & 0.8792 & 0.9104 & 0.7029 & 0.8792 & 0.9104 & 0.7029 & 0.8792 & 0.9104 & 0.7029 & 0.8792 & 0.9104 & 0.7029 \\
$\text{HyperIQA}_{LC}$ & 0.6709 & 0.7304 & 0.4868 & 0.7740 & 0.8107 & 0.5823 & 0.8014 & 0.8448 & 0.6118 & 0.8409 & 0.8752 & 0.6532 \\
GDBC-$\text{HyperIQA}_{LA}$ & 0.8012 & 0.8452 & 0.6109 & 0.8345 & 0.8740 & 0.6474 & 0.8425 & 0.8788 & 0.6563 & 0.8536 & 0.8877 & 0.6717 \\
$\Delta$(\%) & \textbf{+19.4217} & \textbf{+15.7174} & \textbf{+25.4930} & \textbf{+7.8165} & \textbf{+7.8081} & \textbf{+11.1798} & \textbf{+5.1285} & \textbf{+4.0246} & \textbf{+7.2736} & \textbf{+1.5103} & \textbf{+1.4282} & \textbf{+2.8322} \\ \midrule
\bottomrule
\end{tabular}
} 
\end{table*}

\subsubsection{Effectiveness of model bias calibration} 

In this section, we evaluate the LA-MOS prediction performances of different deep IQA models when they are trained with different labels and strategies. As shown in Table \ref{tab:table1}, it is not surprising that all IQA models trained with the LA-MOS achieve better performance than their counterparts trained with the LC-MOS, which have been highlighted by the gray background. These results further confirm the issue of LC-MOS overfitting observed in Fig. \ref{fig:True MSE}. Meanwhile, we can also find that the proposed GDBC efficiently improves the performances of all IQA models when they are trained with LC-MOS, where all $\Delta(\%)$ report positive values across different databases. 

In addition, as shown in each row of Table \ref{tab:table1}, we can find that the performance improvement of GDBC is relatively smaller in CSIQ than in the other databases. There are two possible reasons to account for this result. Firstly, the CSIQ database contains the smallest LC-MOS bias as shown in Table \ref{tab:calibrated-MOS}, which limits the improvement space of GDBC. Secondly, in comparison with the authentic distortion databases (i.e., KONIQ and LIVEC), the diversity of the CSIQ database is limited which reduces the discrepancy between the training and testing sets and the potential overfitting risk. When we look at each column of Table \ref{tab:table1}, it is seen that the performance improvement of GDBC changes significantly across different IQA models, where the DBCNN and HyperIQA obtain higher $\Delta(\%)$. A possible reason lies in the bias propagation. Actually, unlike the plain single-branch architecture in ResNet and NIMA, DBCNN and HyperIQA are composed of multiple branches, which are beneficial for capturing more comprehensive quality-aware features. However, when training with LC-MOS, the subjective bias would easily propagate between different branches and enlarge the misleading effect of LC-MOS. In these cases, the superiority of GDBC would become more significant.

\begin{table*}[] 
\centering
\caption{Performances of different methods with different annotation numbers. $LA$ and $LC$ indicate training using LA-MOS and LC-MOS, respectively.}
\label{tab:table3}
\resizebox{2\columnwidth}{!}{
    \setlength{\tabcolsep}{1pt}
\begin{tabular}{l|rrr|rrr|rrr|rrr}
\toprule 
\toprule 
\multirow{2}{*}{Methods} & \multicolumn{3}{c|}{$M=1$} & \multicolumn{3}{c|}{$M=2$} & \multicolumn{3}{c|}{$M=4$} & \multicolumn{3}{c}{$M=8$} \\
 & SRCC $\uparrow$ & PLCC $\uparrow$ & KRCC $\uparrow$ & SRCC $\uparrow$ & PLCC $\uparrow$ & KRCC $\uparrow$ & SRCC $\uparrow$ & PLCC $\uparrow$ & KRCC $\uparrow$ & SRCC $\uparrow$ & PLCC $\uparrow$ & KRCC $\uparrow$ \\ \midrule
\rowcolor[gray]{.8}$\text{ResNet}_{LA}$ & 0.8985 & 0.9214 & 0.7290 & 0.8985 & 0.9214 & 0.7290 & 0.8985 & 0.9214 & 0.7290 & 0.8985 & 0.9214 & 0.7290 \\
$\text{ResNet}_{LC}$ & 0.7905 & 0.8361 & 0.5990 & 0.8445 & 0.8784 & 0.6580 & 0.8654 & 0.8980 & 0.6854 & 0.8779 & 0.9063 & 0.7014 \\
GDBC-$\text{ResNet}_{LC}$ & 0.8294 & 0.8659 & 0.6425 & 0.8543 & 0.8869 & 0.6706 & 0.8738 & 0.9022 & 0.6948 & 0.8809 & 0.9074 & 0.7051 \\
$\Delta$(\%) & \textbf{+4.9209} & \textbf{+3.5642} & \textbf{+7.2621} & \textbf{+1.1604} & \textbf{+0.9677} & \textbf{+1.9149} & \textbf{+0.9706} & \textbf{+0.4677} & \textbf{+1.3715} & \textbf{+0.3417} & \textbf{+0.1214} & \textbf{+0.5275} \\ \midrule
\rowcolor[gray]{.8}$\text{NIMA}_{LA}$ & 0.8981 & 0.9179 & 0.7287 & 0.8981 & 0.9179 & 0.7287 & 0.8981 & 0.9179 & 0.7287 & 0.8981 & 0.9179 & 0.7287 \\
$\text{NIMA}_{LC}$ & 0.7551 & 0.7363 & 0.5626 & 0.8445 & 0.8795 & 0.6593 & 0.8744 & 0.9012 & 0.6951 & 0.8843 & 0.9062 & 0.7091 \\
GDBC-$\text{NIMA}_{LC}$ & 0.8216 & 0.8519 & 0.6357 & 0.8606 & 0.8915 & 0.6789 & 0.8797 & 0.9038 & 0.7021 & 0.8869 & 0.9064 & 0.7130 \\
$\Delta$(\%) & \textbf{+8.8068} & \textbf{+15.7001} & \textbf{+12.9932} & \textbf{+1.9065} & \textbf{+1.3644} & \textbf{+2.9728} & \textbf{+0.6061} & \textbf{+0.2885} & \textbf{+1.0070} & \textbf{+0.2940} & \textbf{+0.0221} & \textbf{+0.5500} \\ \midrule
\rowcolor[gray]{.8}$\text{DBCNN}_{LA}$ & 0.8899 & 0.9083 & 0.7161 & 0.8899 & 0.9083 & 0.7161 & 0.8899 & 0.9083 & 0.7161 & 0.8899 & 0.9083 & 0.7161 \\
$\text{DBCNN}_{LC}$ & 0.7087 & 0.7474 & 0.5189 & 0.8038 & 0.8436 & 0.6135 & 0.8453 & 0.8751 & 0.6574 & 0.8577 & 0.8857 & 0.6740 \\
GDBC-$\text{DBCNN}_{LC}$ & 0.8053 & 0.8424 & 0.6137 & 0.8418 & 0.8728 & 0.6549 & 0.8631 & 0.8877 & 0.6800 & 0.8707 & 0.8927 & 0.6902 \\
$\Delta$(\%) & \textbf{+13.6306} & \textbf{+12.7107} & \textbf{+18.2694} & \textbf{+4.7275} & \textbf{+3.4614} & \textbf{+6.7482} & \textbf{+2.1058} & \textbf{+1.4398} & \textbf{+3.4378} & \textbf{+1.5157} & \textbf{+0.7903} & \textbf{+2.4036} \\ \midrule
\rowcolor[gray]{.8}$\text{HyperIQA}_{LA}$ & 0.8792 & 0.9104 & 0.7029 & 0.8792 & 0.9104 & 0.7029 & 0.8792 & 0.9104 & 0.7029 & 0.8792 & 0.9104 & 0.7029 \\
$\text{HyperIQA}_{LC}$ & 0.6709 & 0.7304 & 0.4868 & 0.7322 & 0.7890 & 0.5452 & 0.8114 & 0.8569 & 0.6209 & 0.8421 & 0.8825 & 0.6554 \\
GDBC-$\text{HyperIQA}_{LC}$ & 0.8012 & 0.8452 & 0.6109 & 0.8275 & 0.8670 & 0.6399 & 0.8342 & 0.8535 & 0.6485 & 0.8567 & 0.8845 & 0.6733 \\
$\Delta$(\%) & \textbf{+19.4217} & \textbf{+15.7174} & \textbf{+25.4930} & \textbf{+13.0156} & \textbf{+9.8859} & \textbf{+17.3698} & \textbf{+2.8100} & \textbf{+-0.3968} & \textbf{+4.4452} & \textbf{+1.7334} & \textbf{+0.2238} & \textbf{+2.7269} \\  
\midrule    
\bottomrule
\end{tabular}
} 
\end{table*}
\subsection{Robustness toward different subjective bias settings}
Besides different IQA models and databases, we also investigate the robustness of the proposed method toward different subjective bias settings by changing the bias rate and annotation number. More specifically, a higher bias rate and smaller annotation number would cause greater subjective bias. In this section, all experiments are conducted on the KONIQ database. 

\subsubsection{Different bias rates}
Focusing on the challenging cases, we first set the annotation number to the smallest value, i.e., $M=1$, and investigate the performances of different IQA models learned with LC-MOS under the following bias rates $\eta=\{100\%, 80\%, 60\%, $ $40\%\}$. As shown in Table \ref{tab:table2}, the performances of all IQA models gradually increase with a decreasing $\eta$. By contrast, the performance improvement $\Delta(\%)$ keeps falling with the decreasing $\eta$. Since the impact of subjective bias reduces with a decreasing bias rate, the overfitting risk toward LC-MOS would also reduce, which may limit the room for IQA models' improvement from GDBC. Even so, the proposed GDBC improves the LA-MOS prediction accuracy for all IQA models across different bias rates, whose $\Delta(\%)$ values are all positive. This verifies that the proposed GDBC is robust to the variations of bias rates.

\subsubsection{Different annotation numbers}
On the other hand, we set the bias rate to the highest value, i.e., $\eta=100\%$, and investigate the performances of different IQA models with the following annotation numbers $M=\{1, 2, 4, 8\}$. When $M$ increases, the LC-MOS would keep approaching LA-MOS, which also reduces the impact of subjective bias. Similarly, in Table \ref{tab:table3}, we can find that all IQA models' performances increase as the annotation number goes up. Meanwhile, the superiority of the proposed GDBC would keep shrinking. But, our method still achieves positive $\Delta(\%)$ across all annotation numbers in this investigation, which verifies the robustness of GDBC to the variations of annotation numbers.

\subsection{Parameter analysis}
In this section, we further investigate the impact of the parameters $\alpha$ and $\epsilon$ in Eq. \ref{b}, which control the updating intensity and frequency of $\mu_{z_i}$ for our GDBC. More specifically, a larger $\alpha$ would reduce the updating intensity and tend to keep the $\mu_{z_i}$ unchanged in each iteration. Similarly, a larger $\epsilon$ would reduce the updating frequency of $\mu_{z_i}$ in the whole training process and vice verse. All experiments are conducted on the KONIQ database under the most challenging setting, i.e., $\eta=100\%$ and $M=1$.

\begin{figure}  
    \centering
    \includegraphics[width=\linewidth]{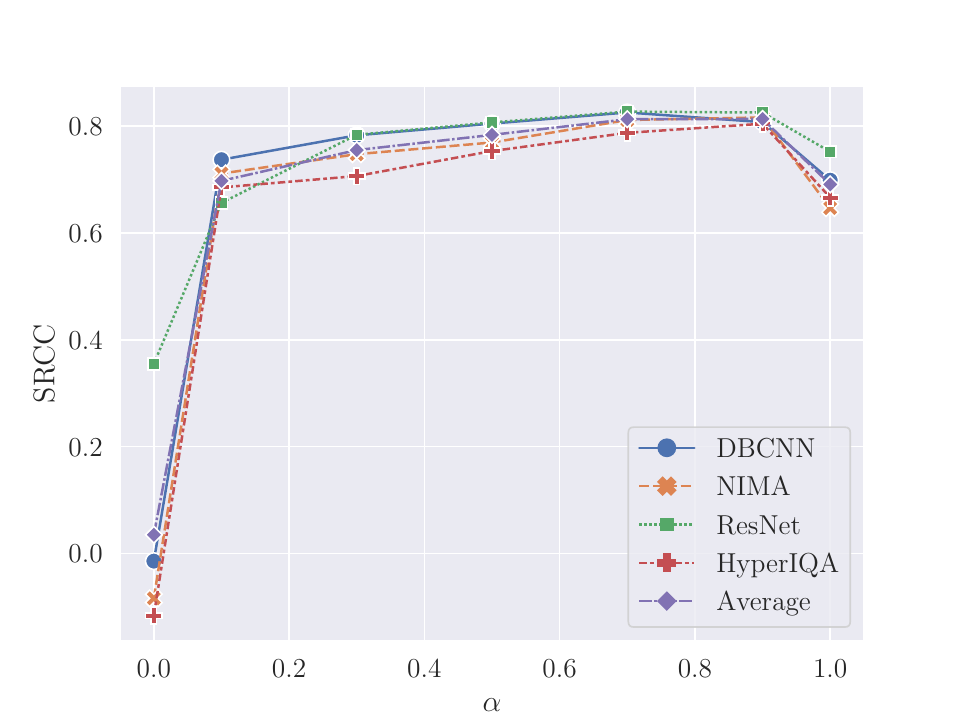}  
    \caption{The performance variation of GDBC with different $\alpha$ on various IQA models.}
    \label{fig:alpha}
\end{figure} 

Referring to Eq. \ref{eq-update}, we can infer that $\alpha$ ranges from 0 to 1. Without loss of generality, we investigate the performance variation of GDBC under seven different $\alpha$ values, i.e., $\{0.0, 0.1, 0.3, 0.5, 0.7, 0.9, 1\}$. As shown in Fig. \ref{fig:alpha}, the performances of all IQA models undergo a process of up and down when $\alpha$ gradually grows from 0 to 1. Actually, too large $\alpha$ would reduce the updating intensity and result in underadjustment toward $\mu_{z_i}$. By contrast, too small $\alpha$ would increase the updating intensity and may result in overadjustment for $\mu_{z_i}$. Although different IQA models prefer different $\alpha$ settings, their performance changes become very slight when $\alpha$ ranges from 0.5 to 0.9. In view of the average performance of all IQA models, we experimentally set $\alpha$ to 0.9 in the proposed method. 

\begin{figure}  
    \centering
    \includegraphics[width=1.0\linewidth]{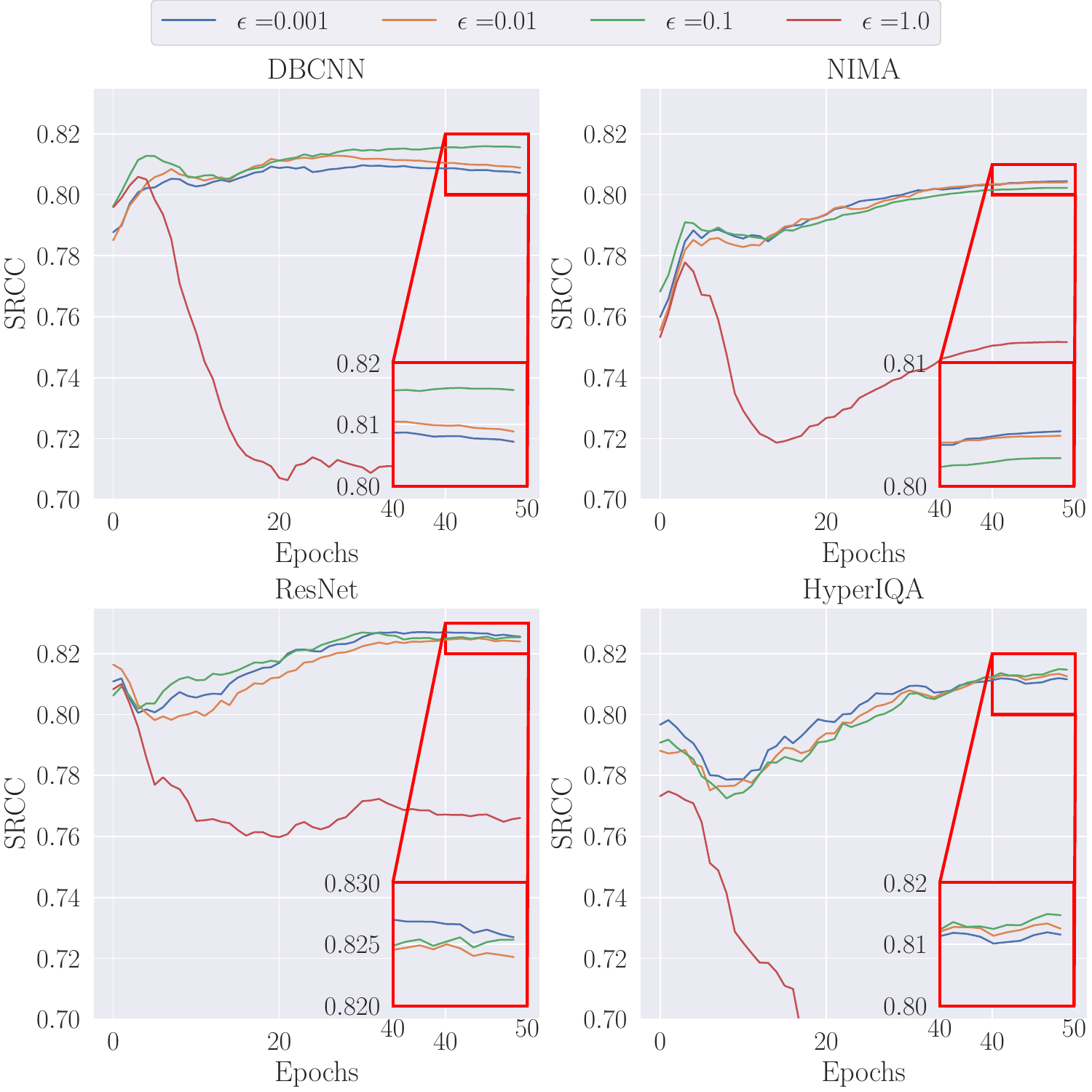}   
    \caption{The performance variation of GDBC with different $\epsilon$ on various IQA models.}
    \label{fig:sth}
\end{figure}

In addition, we also investigate the performance variations of GDBC under different $\epsilon$ values, i.e., $\{0.001, 0.01, 0.1, 1.0\}$. As shown in Fig. \ref{fig:sth}, when $\epsilon$ grows from 0.001 to 0.1, the performances of GDBC are insensitive to different threshold settings, whose curves are very close to each other across all IQA models and the SRCC fluctuations are within 0.02. However, when $\epsilon$ reaches 1, the performance of GDBC significantly drops across all IQA models. As mentioned above, too large $\epsilon$ tends to disable the updating of $\mu_{z_i}$ and degrade GDBC to a plain model, which directly learns from the noisy LC-MOS and suffers their misleading. Based on these investigations, we experimentally set $\epsilon$ to 0.01 in our GDBC. 

\subsection{Comparison with separate subjective bias calibration and model bias calibration methods}

Besides the effectiveness and robustness validation, we further compare our alternating optimization based dual-bias calibration model with some separate subjective bias calibration and model bias calibration methods. On the one hand, many standardized label screening models have been developed to recover high-quality MOS from multiple noisy human annotations, which focus on the subjective bias calibration, such as the subject rejection (SR) model in ITU-R BT.500~\cite{bt2020methodologies}, the maximum likelihood estimation (MLE) model in ITU-T P.910~\cite{installationsitu}, and the subject bias removal (SBR) model in ITU-T P.913~\cite{union2016methods}. In view of the label screening models' request for multiple annotations, we simply the most challenging case of LC-MOS learning to the following setting, i.e., $\eta=100\%$ and $M=2$, and the repetition number of subjective test is set to 1. On the other hand, regarding the subjective bias as the noise, recent noisy label learning models could also be applicable to the model bias calibration in our LC-MOS learning task, such as the generalized cross entropy (GCE) loss \cite{zhang2018generalized} and symmetric cross entropy (SCE) loss based \cite{wang2019symmetric} methods. It is noted that these representative noisy label learning methods, such as GCE and SCE, were originally developed for multiple output based classification networks, which is hard to apply to most single output based regression networks in IQA. For compatibility, we evaluate all bias calibration models on a multiple output based IQA model NIMA~\cite{talebi2018nima}. Meanwhile, in view of the availability of raw human annotations, we conduct all experiments on the VCL database in this section.    

As shown in Table \ref{tab:tablef}, both the label screening and noisy label learning methods deteriorate the performance of NIMA when LC-MOS is used for training. Only the proposed GDBC improves the IQA model learning. Firstly, unlike our iterative optimization strategy, existing label screening methods like SR, MLE, and BR adopt a one-stop post-processing paradigm, which can not identify or even suppress the error in calibrating the human annotations. When the available annotations are very few, the calibration error easily becomes considerable and further interferes with the IQA model learning. Secondly, to avoid overfitting issues, existing noisy label learning methods like GCE and SCE focus on suppressing the gradient backpropagation, which slows down the rate of convergence and easily causes underfitting in turn. These reported results further verify the superiority and necessity of jointly calibrating the subjective bias and model bias.

\begin{table}[]
\centering
\caption{Comparison results with the representative label screening and noisy label learning methods}
\label{tab:tablef}
  \scalebox{1}{
\begin{tabular}{l|c|c|c}
\toprule 
\midrule 
Methods & SRCC & PLCC & KRCC \\ \midrule
NIMA & 0.9142 & 0.8770 & 0.7311 \\ \midrule
SR-NIMA & 0.8976 & 0.8419 & 0.7061 \\ \midrule
MLE-NIMA & 0.8808 & 0.8805 & 0.6921 \\ \midrule
BR-NIMA & 0.9090 & 0.8742 & 0.7303 \\ \midrule 
GCE-NIMA & 0.7788 & 0.7608 & 0.6120 \\ \midrule
SCE-NIMA & 0.8772 & 0.8587 & 0.7237 \\ \midrule
GDBC-NIMA & \textbf{0.9233} & \textbf{0.9053} & \textbf{0.7500}  \\ \midrule
\bottomrule
\end{tabular}}
\end{table}

\section{Conclusion}

In this paper, we explore a new challenge to learn robust IQA models from noisy low-cost MOS (LC-MOS), which requires very few opinion scores for each image. By jointly inferring the subjective bias and model bias, we develop a plug-and-play gated dual-bias calibration (GDBC) module, which enforces the IQA model learned from LC-MOS to approach the unbiased estimation of labor-abundant MOS (LA-MOS). Extensive experiments on four popular IQA databases and four representative deep IQA models verify the effectiveness of the proposed method, which significantly outperforms the IQA models directly learned from LC-MOS and even achieves comparable performance with respect to the IQA models learned from the expensive and time-consuming LA-MOS. Meanwhile, we also verify the superiority of the proposed method over the existing label screening and noisy label learning methods.

\bibliographystyle{IEEEtran}
\bibliography{new}
\end{document}